\pdfoutput=1

\documentclass[11pt]{article}

\usepackage{acl}

\usepackage{times}
\usepackage{latexsym}
\usepackage{amssymb}

\usepackage[T1]{fontenc}

\usepackage[utf8]{inputenc}

\usepackage{microtype}
\usepackage{amsmath}

\usepackage{booktabs}
\usepackage{algorithm}
\usepackage{algorithmic}
\usepackage{multirow}
\usepackage[most]{tcolorbox}
\tcbuselibrary{skins, breakable}
\usepackage{enumitem}
\usepackage{colortbl}
\usepackage{subcaption}

\usepackage{soul}
\usepackage{xcolor}
\usepackage{tikz}
\usetikzlibrary{tikzmark}
\definecolor{lightgreen}{RGB}{197, 224, 180}
\sethlcolor{lightgreen}
\usepackage{listings}

\makeatletter
\newcommand*\myfontsize{%
  \@setfontsize\myfontsize{8}{8}%
}
\newcommand{\mytextbox}[2]{\tikzmarknode[draw=#1,thick,inner sep=2pt,outer sep=2pt]{test}{\myfontsize #2}}

\newcommand{\symboletongyi}{\raisebox{0pt}{~\includegraphics[scale=0.012]{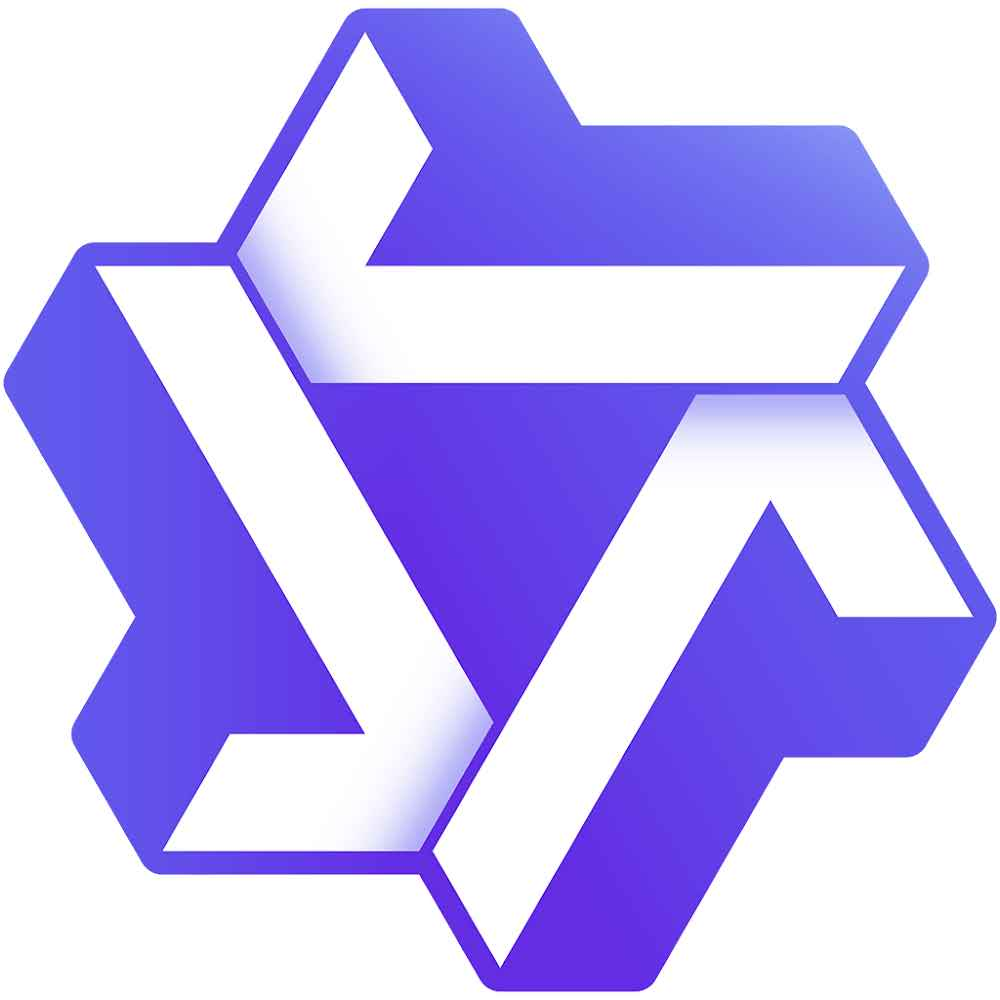}}~}

\definecolor{myred}{rgb}{0.7, 0.3, 0.0}
\definecolor{myblue}{rgb}{0.2, 0.3, 0.6}
\definecolor{mygreen}{HTML}{008000}
\definecolor{mypurple}{rgb}{0.5, 0.0, 0.8}

\newcommand{\analysis}{\mytextbox{mygreen}{\textbf{\textcolor{mygreen}{<analysis>}}}}

\newcommand{\evidence}{\mytextbox{myred}{\textbf{\textcolor{myred}{<evidence>}}}}

\newcommand{\reasoning}{\mytextbox{myblue}{\textbf{\textcolor{myblue}{<reasoning>}}}}

\newcommand{\answer}{\mytextbox{mypurple}{\textbf{\textcolor{mypurple}{<answer>}}}}

\usepackage{inconsolata}

\usepackage{graphicx}
\usepackage{forest}

\usepackage{longtable}

\title{Evidence-Augmented Policy Optimization with Reward Co-Evolution for Long-Context Reasoning}

\author{
\textbf{Xin Guan}\thanks{Work done during the author's internship at Tongyi Lab.}, 
\textbf{Zijian Li}, 
\textbf{Shen Huang}, 
\textbf{Pengjun Xie}, \\
\textbf{Jingren Zhou},
\textbf{Jiuxin Cao}\\
Tongyi Lab\symboletongyi, Alibaba Group
}

\begin{document}
\maketitle

\begin{abstract}
While Reinforcement Learning (RL) has advanced LLM reasoning, applying it to long-context scenarios is hindered by sparsity of outcome rewards. This limitation fails to penalize ungrounded ``lucky guesses,'' leaving the critical process of needle-in-a-haystack evidence retrieval largely unsupervised. To address this, we propose \textbf{EAPO} (\textbf{E}vidence-\textbf{A}ugmented \textbf{P}olicy \textbf{O}ptimization). We first establish the Evidence-Augmented Reasoning paradigm, validating via Tree-Structured Evidence Sampling that precise evidence extraction is the decisive bottleneck for long-context reasoning. Guided by this insight, EAPO introduces a specialized RL algorithm where a reward model computes a Group-Relative Evidence Reward, providing dense process supervision to explicitly improve evidence quality. To sustain accurate supervision throughout training, we further incorporate an Adaptive Reward-Policy Co-Evolution mechanism. This mechanism iteratively refines the reward model using outcome-consistent rollouts, sharpening its discriminative capability to ensure precise process guidance. Comprehensive evaluations across eight benchmarks demonstrate that EAPO significantly enhances long-context reasoning performance compared to SOTA baselines.
\end{abstract}

\begin{figure}[t!] 
    \centering
    \includegraphics[width=0.99 \columnwidth]{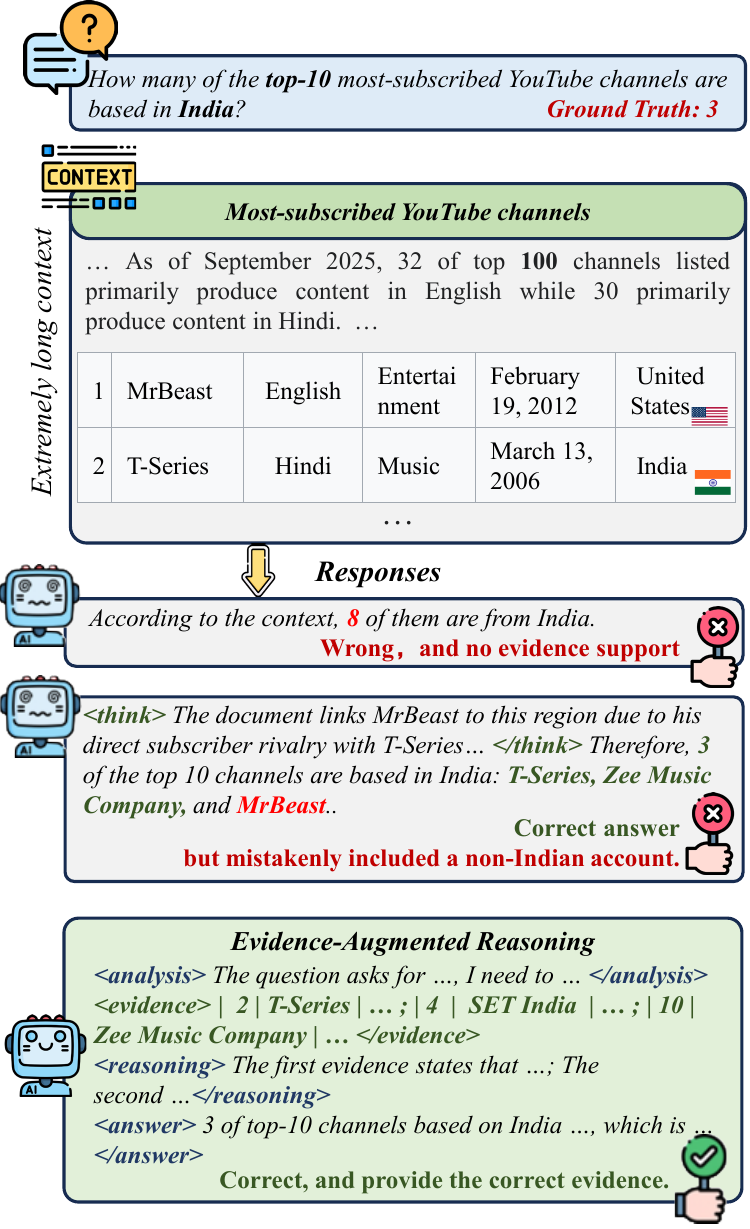}
    \caption{Challenges in reasoning over extremely long contexts. The examples illustrate critical failure modes, including answers with no evidence support, incomplete or error retrieval. Our work explicitly targets this bottleneck to ensure high-fidelity evidence extraction.}
    \label{figure-intro}
\end{figure}

\section{Introduction}
The development of Large Language Models (LLMs) has profoundly revolutionized Natural Language Processing (NLP) \cite{DBLP:journals/corr/abs-2501-09686}. Models such as GPT \cite{DBLP:journals/corr/gpt-oss} and DeepSeek-R1 \cite{DBLP:journals/corr/dpsk} have propelled significant advancements in domains ranging from complex dialogue to summarization. However, as LLMs evolve into autonomous agents required to interact with dynamic environments, the demand for processing voluminous information has escalated. This imposes a critical challenge: as context length increases, model performance often suffers from severe degradation due to the ``lost in the middle'' phenomenon \cite{DBLP:journals/tacl/LiuLHPBPL24}. In these high-noise scenarios, models frequently exhibit specific failure modes illustrated in Figure \ref{figure-intro}: they may generate answers with \textbf{no evidence support}, rely on \textbf{fragmentary retrieval}, or make \textbf{erroneous citations}, failing to discriminate ``needle-in-a-haystack'' facts from irrelevant distractors.

Reinforcement Learning (RL) \cite{DBLP:journals/corr/abs-2509-08827} has emerged as a powerful tool for unlocking intrinsic reasoning capabilities. However, existing approaches in Long-Context Reasoning (LCR) predominantly rely on \textbf{outcome-based rewards} \cite{DBLP:journals/corr/abs-2505-17667, DBLP:journals/corr/abs-2510-19363}, guiding optimization solely based on the correctness of the final answer. This reliance exacerbates reward sparsity and may introduce noise: a model may correctly guess an answer via a flawed reasoning path \cite{shao2025deepseekmath}. For instance, as shown in Figure \ref{figure-intro}, a model might correctly answer ``3 channels'' but cite an irrelevant entity (e.g., \textit{MrBeast}) instead of the correct Indian channel. Such "shortcuts" satisfy the binary outcome reward but fail to generalize, leaving the underlying retrieval failure unaddressed.

To resolve these issues, we argue for a shift from result-oriented to evidence-augmented process-oriented supervision for LCR. To operationalize this, we first establish the Evidence-Augmented Reasoning (EAR) paradigm. By strictly enforcing a workflow where explicit evidence extraction precedes reasoning execution, EAR inherently resolves the problem of ungrounded responses, ensuring that no conclusion is derived without explicit citation. Building on this structured foundation, we conduct a preliminary study via Tree-Structured Evidence Sampling, which employs BFS to explore distinct evidence trajectories, evaluated via both semantic similarity and fine-grained LLM judgments. This analysis empirically validates our core hypothesis: securing high-quality evidence is the decisive prerequisite for unlocking superior performance in LCR.

Building on these insights, we propose \textbf{EAPO} (\textbf{E}vidence-\textbf{A}ugmented \textbf{P}olicy \textbf{O}ptimization). EAPO is designed to internalize the benefits of precise evidence discovery directly into the model's policy. It introduces a dense, group-relative evidence reward to provide fine-grained supervision at the process level, ensuring the model is rewarded not just for being right, but for being right for the high-quality evidence. 
However, relying on a static reward model carries risks, as the policy improves, a fixed evaluator may fail to distinguish subtle quality differences. To address this, we further incorporate an Adaptive Reward-Policy Co-Evolution mechanism. This creates a self-reinforcing cycle where the reward model is iteratively refined using high-confidence, outcome-consistent rollouts generated by the policy itself. This continuous alignment sharpens the reward model's discriminative capability, enabling it to assess complex evidence with increasing precision and dynamically synchronize its supervision signal with the evolving policy.

The core contributions of our paper are:
\begin{itemize}
    \item We establish the EAR paradigm to enforce explicit evidence extraction prior to reasoning, and empirically identify this extraction step as the decisive bottleneck through Tree-Structured Evidence Sampling.
    \item We introduce the \textbf{EAPO} framework, shifting optimization from sparse outcome rewards to dense, evidence-augmented supervision. By leveraging a reward model to compute a group-relative evidence reward, EAPO guides the model to extract high-quality evidence, thereby enhancing reasoning capabilities.
    \item We design an Adaptive Reward-Policy Co-Evolution mechanism to establish a self-reinforcing loop. By iteratively refining the reward model with outcome-consistent rollouts, this mechanism sustains high-fidelity supervision.
\end{itemize}

\section{Empirical Motivation: The Primacy of Evidence}
Before detailing our reinforcement learning framework, we first establish empirical foundations of our approach. We introduce the Evidence-Augmented Reasoning (EAR) paradigm to structure the long-context workflow and utilize a Tree-Structured Evidence Sampling to validate a critical hypothesis that \textit{evidence extraction is the decisive bottleneck in long-context reasoning}.

\subsection{Evidence-Augmented Reasoning}
We propose the \textbf{Evidence-Augmented Reasoning (EAR)} paradigm, as shown in Figure \ref{figure-intro}, which explicitly decouples the retrieval of information from its logical manipulation. EAR enforces a strict four-stage workflow:
1) \textbf{Task Analysis \analysis }: Deconstructing the user's query constraints.
2) \textbf{Evidence Extraction \evidence}:  Locating and quoting relevant text segments verbatim from the context.
3) \textbf{Reasoning Execution \reasoning}: Deriving conclusions based exclusively on the extracted evidence.
4) \textbf{Answer Formulation \answer}: Synthesizing the final response.
This structural decomposition, operationalized via special tokens, not only focuses attention but crucially exposes the intermediate evidence state for direct supervision.

\begin{figure}[t]
    \centering
    \includegraphics[width=0.99 \columnwidth]{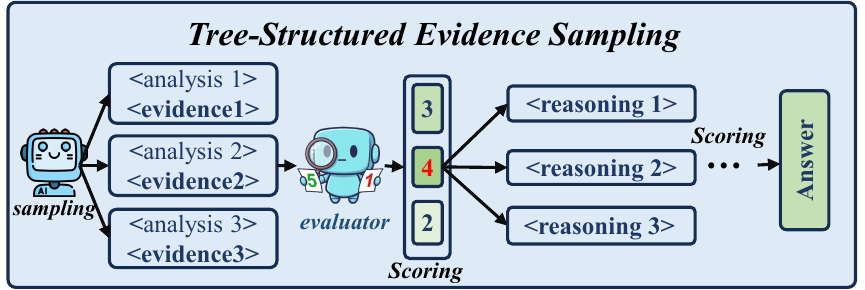}
   \caption{Overview of Tree-Structured Evidence Sampling. The process samples diverse evidence and reasoning paths, evaluates their scores, and derives the final answer.}
    \label{fig2}
\end{figure}

\subsection{Preliminary Experiment on Evidence}
\paragraph{Tree-Structured Evidence Sampling.}
To quantify the critical role of evidence quality in determining final reasoning outcomes, we design a systematic exploration strategy modeled as a decision tree. As shown in Figure \ref{fig2}, we employ Breadth-First Search (BFS) to sample $k$ distinct evidence trajectories for a given query and context, creating a diverse search space. To efficiently value these intermediate nodes without redundant full-context processing, we apply a Query-Centric Node Evaluation that scores candidates based strictly on their utility relative to the query. We investigate two specific valuation paradigms to identify the better evidence:
\begin{itemize}
    \item \textbf{Semantic Similarity Assessment (SSA):} Using a reranker (Qwen3-8B-Reranker \cite{DBLP:journals/corr/abs-2506-05176} ) to measure the coarse-grained embedding alignment between the query and the extracted evidence.
    \item \textbf{LLM-as-a-Judge (LLM-E/R):} Leveraging an LLM to perform fine-grained evaluation on the utility of the extracted evidence (LLM-E) or the coherence of the reasoning (LLM-R) (Prompts in Appendix \ref{evidence prompt}).
\end{itemize}

\paragraph{Experiment Setting} 
\label{pre-exp}

\begin{figure}[t]
    \centering
    \includegraphics[width=0.99 \columnwidth]{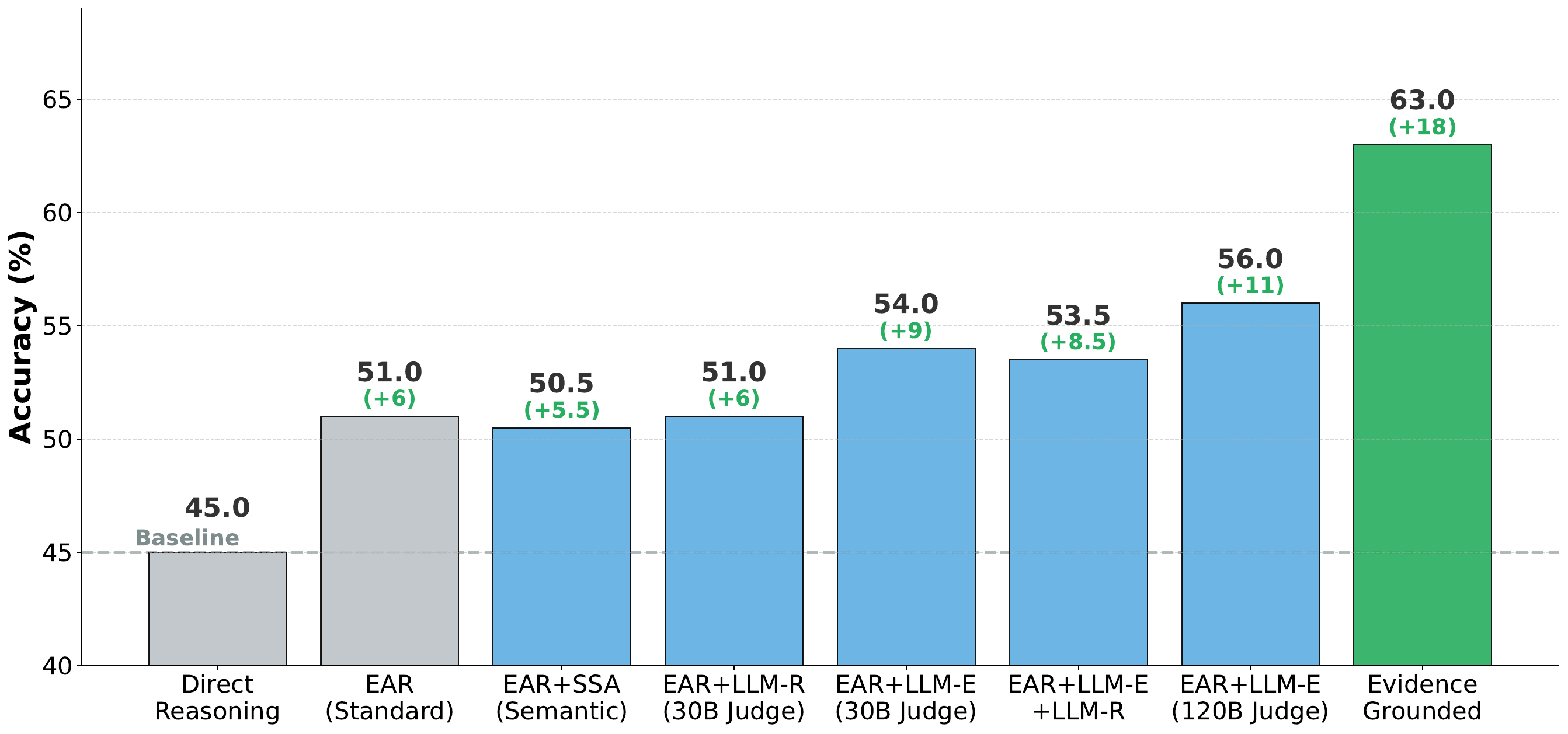}
    \caption{Performance analysis of Qwen3-30B-A3B-Instruct on Musique in LongBench. We compare Direct Reasoning, EAR variants with different evaluators, and an Oracle upper bound (\textit{Evidence-Grounded}).}
    \label{fig:pilot_study}
\end{figure}

To quantify the factors constraining long-context performance, we conduct a controlled analysis on the Musique from LongBench benchmark \cite{DBLP:conf/acl/BaiLZL0HDLZHDTL24} using Qwen3-30B-A3B-Instruct \cite{DBLP:journals/corr/abs-2505-09388} as the base model. To establish an empirical performance ceiling, we construct an Evidence-Grounded (Oracle) baseline, where high-fidelity evidence is extracted by a superior model (\texttt{gemini-2.5-pro} \cite{gemini2.5}) utilizing both the query and ground-truth answer, and then fed to the base model for inference. We employ Tree-Structured Sampling with a beam width of $k=3$. As illustrated in Figure \ref{fig:pilot_study}, our analysis yields the following insights:

\paragraph{Evidence Extraction is the Decisive Bottleneck.}
Two observations isolate evidence retrieval as the primary bottleneck. First, providing high-fidelity evidence (Oracle) drastically elevates performance to 63.0\%. Second, optimizing solely the reasoning step (EAR+LLM-R) yields no improvement over standard EAR. Moreover, adding a reasoning judge atop evidence evaluation (EAR+LLM-E+LLM-R) has a negligible effect, causing a slight performance drop compared to LLM-E alone. We attribute this to restricted reasoning diversity: once evidence is fixed, subsequent reasoning paths become highly homogeneous (>96\% semantic consistency; see Appendix \ref{sec:reasoning_diversity}), rendering LLM-R redundant. This implies that once high-quality evidence is secured, inherent reasoning capabilities suffice; finding the evidence is the true bottleneck.

\begin{figure*}[t!]
    \centering
    \includegraphics[width=0.99 \textwidth]{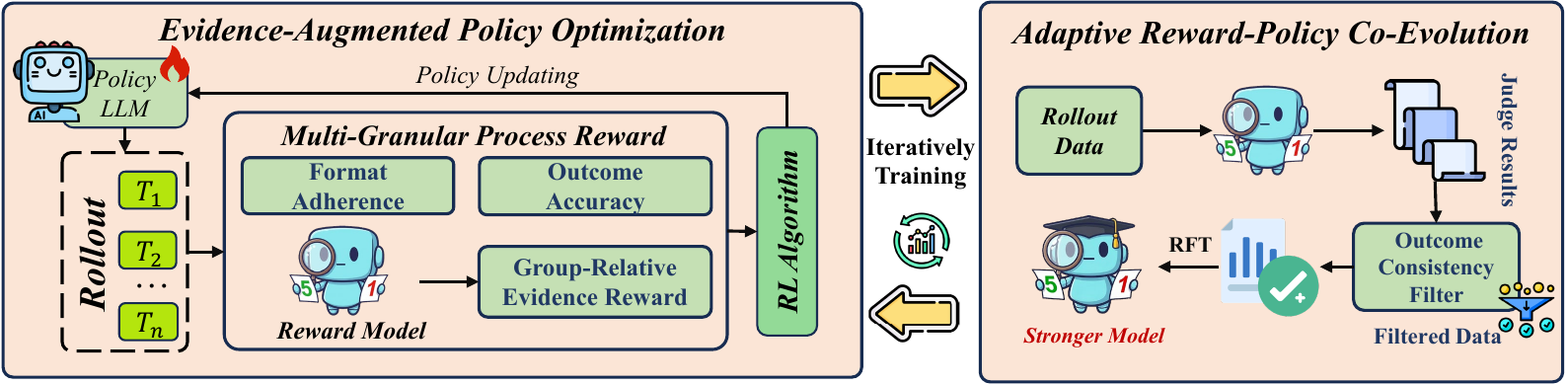} 
    \caption{Overview of EAPO. \textbf{Left:} The Evidence-Augmented Policy Optimization phase, utilizing dense evidence scores to guide the high-quality evidence extraction process. \textbf{Right:} The Adaptive Reward-Policy Co-Evolution cycle, which iteratively refines the reward model to sharpen its discriminative capability for accurate supervision.}
    \label{fig_eapo}
\end{figure*}

\paragraph{Superior Process Supervision Drives Greater Performance Gains.}
Our analysis reveals a direct correlation between the fidelity of process supervision and the magnitude of performance improvement. While SSA offers limited guidance (+5.5\%), employing an LLM-based judge to provide process supervision significantly improves performance (+9.0\%). Furthermore, elevating the supervision quality with a stronger gpt-oss-120B evaluator yields even greater gains (+11.0\%).

\section{Evidence-Augmented Policy Optimization}

Building on the empirical findings in Section \ref{pre-exp}, we propose Evidence-Augmented Policy Optimization (EAPO). Guided by the insight that evidence extraction is the decisive bottleneck, EAPO shifts supervision from sparse outcome signals to dense, process-oriented guidance. Furthermore, to leverage the critical correlation between evaluator strength and performance, we incorporate an Adaptive Reward-Policy Co-Evolution mechanism. This establishes a self-reinforcing loop where the reward model's discriminative power is iteratively sharpened in tandem with the policy. The framework's overview is depicted in Figure \ref{fig_eapo}.

\subsection{Preliminaries}
\paragraph{Group Relative Policy Optimization (GRPO).}
We adopt Group Relative Policy Optimization (GRPO) \cite{DBLP:journals/corr/abs-2402-03300} as the core reinforcement learning algorithm. 
Formally, for each query $q$ sampled from the dataset distribution $P(Q)$, we generate a group of $G$ outputs $\{o_1, ..., o_G\}$ utilizing the old policy $\pi_{\theta_{old}}$. The optimization objective is to maximize the following surrogate loss function:

\begin{equation}
    \begin{split}
        \mathcal{L}_{GRPO}(\theta) = \mathbb{E}_{q \sim P(Q), \{o_i\} \sim \pi_{\theta_{old}}} \bigg[ \frac{1}{G} \sum_{i=1}^{G} \min \Big( \\
        r_i(\theta) \hat{A}_i, \quad \text{clip}\big(r_i(\theta), 1-\epsilon_{low}, 1+\epsilon_{high}\big) \hat{A}_i \Big) \bigg]
    \end{split}
\end{equation}

where $r_i(\theta) = \frac{\pi_\theta(o_i|q)}{\pi_{\theta_{old}}(o_i|q)}$ denotes the probability ratio between the current and old policies. The hyperparameters $\epsilon_{low}$ and $\epsilon_{high}$ constrain the policy update step size, set to 0.2 and 0.28 respectively. 
$\hat{A}_i$ represents the advantage value for the $i$-th output, which is derived from the standardized total reward: $\hat{A}_i = \frac{R_{total}^{(i)} - \text{mean}(R_{total})}{\text{std}(R_{total})}$. $R_{total}^{(i)}$ denotes the reward for the $i$-th sample. We define the specific composition of this fine-grained process reward $R_{total}$ in the subsequent section.

\subsection{Multi-Granular Process Reward Mechanism}
EAPO constructs a dense, process reward $R_{total}$. This composite signal is composed of distinct components designed to guide the model through the structured Evidence-Augmented workflow:

\paragraph{Format Adherence ($R_f$).}
To ensure the output conforms to the EAR structure, we verify the presence of specific delimiters (e.g., \texttt{<evidence>}). If the format is correct, we assign $R_f=1$; otherwise, $R_f=0$. If the format is invalid, the evaluation terminates immediately, and the remaining reward components are not calculated.

\paragraph{Group-Relative Evidence Quality ($R_e$).}
This component serves as an intermediate process reward designed to assess the quality of the extracted evidence. Specifically, we extract all evidence segments generated within all rollouts and input this collective set simultaneously into the Reward Model $\theta_{RM}$ for assessment. The RM assigns an integer utility score $v \in \{1, \dots, 5\}$ (Prompt in the Appendix \ref{evidence prompt}). These scores are then normalized within the sampled group to yield $R_e \in [0, 1]$, enabling the policy to prioritize superior evidence through comparative feedback.

\paragraph{Outcome Accuracy ($R_a$).}
Following \cite{DBLP:journals/corr/abs-2505-17667}, we employ an LLM-based evaluator to assess answer correctness. This module assigns a binary reward by verifying if the generated answer is semantically consistent with the ground truth.

\paragraph{Total Reward.}
The composite reward is a weighted linear combination:
\begin{equation}
    R_{total} = \alpha R_f + \beta R_e + \gamma R_a
\end{equation}
where $\alpha, \beta, \gamma$ are hyperparameters balancing structural compliance, evidence quality, and final accuracy.

\subsection{Adaptive Reward-Policy Co-Evolution}
Our preliminary analysis in Section \ref{pre-exp} revealed a critical phenomenon: the efficacy of policy model optimization is positively correlated with the quality of process supervision. To improve the capabilities of the process Reward Model (RM), EAPO incorporates an Adaptive Reward-Policy Co-Evolution mechanism, establishing a dynamic closed-loop where the RM iteratively improves alongside the Policy. This cycle proceeds in two key phases:

\paragraph{Exploratory Rollout and Scoring.}
During the optimization phase, the current policy $\pi_\theta$ generates diverse reasoning trajectories for each query. The current RM evaluates the evidence sets extracted within these rollouts, assigning integer quality scores to guide the immediate policy update.

\paragraph{Outcome-Consistent Rejection Fine-Tuning (RFT).}
To prevent supervision degradation, we employ a strategy inspired by Rejection Fine-Tuning (RFT). Rather than naively training on all correct paths, we apply a strict Outcome Consistency filter to curate a high-fidelity dataset. We retain only those trajectories where the RM's judgment aligns with objective reality. Specifically, instances where high evidence scores correspond to correct answers and low scores to incorrect ones.
These validated High-Confidence data are then used to fine-tune the Reward Model parameters $\theta_{RM}$ via a supervised objective:
\begin{equation}
    \theta_{RM} \leftarrow \arg\min_{\theta} \mathbb{E}_{(q, o, y) \sim \mathcal{D}_{high}} \left[ \mathcal{L}(R_{\theta}(q, o), y) \right]
\end{equation}
where $q$ denotes the input query, $o$ represents the evidence set, and $y$ indicates the target prediction derived from the consistency check.
This process continuously sharpens the RM's discriminative boundary, ensuring that the supervision signal remains accurate and precise as the policy explores increasingly complex reasoning patterns.

By synchronizing the evolution of the reward model with the policy, EAPO creates a self-reinforcing cycle that guides the model toward autonomous, high-fidelity evidence discovery.

\section{Experiments}

\begin{table*}[!t]
    \centering
    \resizebox{0.99 \textwidth}{!}{
    \begin{tabular}{lccccccccc}
    \toprule
    \multirow{2}{*}{Models} & \multicolumn{2}{c}{SEAL} & \multicolumn{3}{c}{LongBench} & \multicolumn{3}{c}{LongBench-v2} & \multirow{2}{*}{AVG} \\
    \cmidrule(lr){2-3} \cmidrule(lr){4-6} \cmidrule(lr){7-9}
     & Seal-0 & Seal-hard & Musique & Hotpotqa & 2wiki & SDQ & MDQ & LSA & \\
    \midrule
    \rowcolor{gray!20} 
    \multicolumn{10}{c}{\textit{Comparison with other LLMs}} \\
    \midrule
    GPT-4o & 19.8 & 39.4 & 60.0 & 78.0 & 79.5 & 47.0 & \underline{50.0} & 44.4 & 52.3 \\
    Gemini-2.0-Flash & 24.3 & 36.6 & 52.5 & 75.0 & 84.5 & 43.1 & 44.1 & 48.2 & 51.0 \\
    Qwen3-Plus & 24.3 & 32.3 & 56.5 & 71.5 & 80.0 & 50.3 & 44.1 & 55.6 & 51.8 \\
    Claude-Sonnet-4 & \underline{42.3} & 56.3 & 60.5 & \textbf{81.5} & 81.0 & 50.3 & 48.0 & \underline{66.7} & \underline{60.8} \\
    GPT-OSS-120B & 39.6 & 55.5 & 64.0 & 80.0 & \textbf{87.5} & 45.7 & 46.1 & 63.0 & 60.2 \\
    QwenLong-32B & 40.5 & \underline{56.7} & \underline{64.5} & 77.5 & 84.5 & 49.0 & \textbf{52.9} & 37.0 & 57.8 \\ 
    \midrule
    \rowcolor{gray!20}
    \multicolumn{10}{c}{\textit{Comparison among Our Models}} \\
    \midrule
    Qwen3-14B & 35.1 & 49.6 & 56.0 & 79.5 & 83.5 & 40.4 & 42.2 & 38.9 & 53.2 \\
    + GRPO  & 35.1 & 51.0 & 57.0 & 80.0 & 83.0 & 40.7 & 42.4 & 40.9 & 53.8 \small{\textcolor{teal}{(+0.6)}} \\
    + EAPO  & 36.0 & 52.5 & 57.0 & \underline{81.0} & 83.5 & 43.0 & 44.1 & 42.9 & 55.0 \small{\textcolor{teal}{(+1.8) }} \\
    \midrule
    Qwen3-30B-A3B-Instruct & 23.4 & 30.3 & 45.0 & 73.0 & 75.0 & 39.7 & 43.1 & 55.6 & 48.1\\
     + GRPO & 31.5 & 40.2 & 51.5 & 75.5 & 85.5 & 41.1 & 43.6 & 55.6 & 53.1 \small{\textcolor{teal}{(+5.0)}} \\
     + EAPO & 33.0 & 44.1 & 59.5 & 77.5 & \textbf{87.5} & 42.4 & 45.1 & 55.6 & 55.6 \small{\textcolor{teal}{(+7.5)}} \\
    \midrule
    Qwen3-30B-A3B-Thinking & 33.3 & 55.5 & 60.0 & 79.0 & 84.5 & 47.0 & 45.1 & 59.3 & 58.0 \\
    + GRPO & 40.4 & 53.8 & 60.5 & 78.5 & 85.0 & 49.7 & 43.1 & 63.0 & 59.2 \small{\textcolor{teal}{(+1.2)}} \\
    + EAPO & \textbf{44.1} & \textbf{57.9 }& \textbf{65.0} & 80.0 & \underline{86.5} & \textbf{51.7} & 49.0 & \textbf{70.4} & \textbf{63.1} \small{\textcolor{teal}{(+5.1)}} \\
    \bottomrule
    \end{tabular}
    }
    \caption{
    Main results on long-context reasoning benchmarks. The best result is highlighted in bold, and the second best result is shown underlined.
    }
    \label{tab:overall}
\end{table*}

\paragraph{Benchmarks}
We evaluate our model on a comprehensive suite of eight long-context reasoning benchmarks, categorized into three types: (1) SEAL-0 and (2) SEAL-Hard from SEAL \cite{DBLP:journals/corr/abs-2506-01062}, which represent challenging reasoning QA tasks \footnote{SEAL provides the Wikipedia URLs for the answers. We employ Jina to extract the page content as the context.}; a collection of multi-hop QA datasets from LongBench-V1 \cite{DBLP:conf/acl/BaiLZL0HDLZHDTL24}, specifically (3) HotpotQA, (4) MuSiQue, and (5) 2WikiMultihopQA; and three diverse tasks from LongBench-V2 \cite{DBLP:conf/acl/BaiTZ0WLCX0D0L25}: (6) Single-Document QA (SDQ), (7) Multi-Document QA (MDQ), and (8) Long Structured-Data QA (LSA).

\paragraph{Baselines} 
We compare our approach against state-of-the-art models divided into three distinct categories:
\begin{itemize}
    \item \textit{Compared LLMs}: High-performing closed-source models including GPT-4o \cite{hurst2024gpt}, Gemini-2.0-Flash \cite{gemini2.0}, Qwen3-Plus, and  Claude-Sonnet-4 \cite{claude}. Leading open weights models such as Qwen3-14B, Qwen3-30B-A3B-Instruct, Qwen3-30B-A3B-Thinking \cite{DBLP:journals/corr/abs-2505-09388}, and GPT-OSS-120B \cite{DBLP:journals/corr/gpt-oss}. \textbf{QwenLong-32B}, the SOTA model for specialized long-context reasoning trained via RL.
    \item \textit{Control Baselines (GRPO)}: To rigorously validate our algorithm, we include variants of our base models trained using \textit{only} the outcome-based reward via standard GRPO. These serve as the primary control group to isolate the specific contribution of our evidence-based process rewards.
\end{itemize}

\paragraph{Evaluation \& Training Details}
We employ \texttt{gpt-4o-2024-11-20} as the judge model to evaluate answer accuracy; the exact evaluation prompt is provided in the Appendix \ref{prompt}. To ensure consistent long-context assessment, we restrict all training and evaluation samples to a maximum context length of 128k tokens.
For training, we utilize three base models to cover diverse scales, architectures, and reasoning patterns: Qwen3-14B (Dense), Qwen3-30B-A3B-Instruct (MoE), and Qwen3-30B-A3B-Thinking \cite{DBLP:journals/corr/abs-2505-09388}. The Reward Model is initialized from the Qwen3-30B-A3B-Thinking. Regarding the reward coefficients, we set the format weight $\alpha=0.1$ following \citet{verl}, while $\beta=0.3$ and $\gamma=0.6$ are determined based on our subsequent hyperparameter analysis. The Reward Model is updated every 20 RL steps using the most recently collected high-quality rollouts. Detailed hyperparameters are listed in the Appendix \ref{appendix:exp-detail}.

\paragraph{Training Dataset}
We construct a composite training dataset totaling 4,664 samples, derived from two sources. First, adopting the long-context construction method of \citet{DBLP:journals/corr/abs-2502-20790}, we process the MuSiQue dataset by padding the context to a range of 32k--128k tokens. Second, we incorporate a Wikipedia-based Mixed QA subset, comprising both structured QA (reasoning over tables) and unstructured QA (free-text extraction). Detailed construction process in Appendix \ref{appendix:exp-detail}. This hybrid composition ensures the model adapts to heterogeneous evidence formats.

\subsection{Main Results}
The overall performance comparisons are presented in Table \ref{tab:overall}. Our analysis yields three significant observations:

\paragraph{Robust Improvements Across Diverse LLMs.}
The proposed EAPO framework demonstrates consistent superiority over base models across all tested architectures.  For Qwen3-30B-Instruct, our method achieves a remarkable average improvement of +7.5\%, boosting the score from 48.1\% to 55.6\%. Even on the highly capable Qwen3-30B-Thinking model, which already possesses advanced reasoning capabilities, EAPO delivers a substantial +5.1\% gain, achieving a score of 63.1\%. This confirms that optimizing the evidence extraction process is universally effective, functioning independently of model size (14B vs. 30B) or architecture type (Dense vs. MoE).

\paragraph{Efficacy of Evidence-Augmented Supervision.}
Comparing outcome-based GRPO with EAPO underscores the critical value of fine-grained guidance. While GRPO yields only marginal gains (e.g., +1.2\% for 30B), EAPO's dense evidence reward drives significant improvements—notably surpassing the GRPO baseline by over 4 points on SEAL-Hard (57.9\% vs. 53.8\%). This confirms that explicit evidence supervision effectively curbs parametric shortcuts, fostering robust and generalizable reasoning.

\paragraph{Competitiveness with Larger and Proprietary Models.}
Remarkably, our EAPO-30B-Thinking model (63.1\%) outperforms significantly larger open-source models like GPT-OSS-120B (60.2\%) and specialized long-context models like QwenLong-32B (57.8\%). Furthermore, it surpasses powerful proprietary models such as Gemini-2.0-Flash (51.0\%), GPT-4o (52.3\%), and Claude-Sonnet-4 (60.8\%) on average, highlighting the efficiency of evidence-augmented optimization.

\section{Analysis Experiment}
Experiments in this section are conducted using the Qwen3-30B-A3B-Thinking model. We adopt this unified setting to rigorously assess the individual contributions of the EAPO framework components.

\subsection{Ablation Study}

\begin{figure}[t]
    \centering
    \includegraphics[width=0.99 \columnwidth]{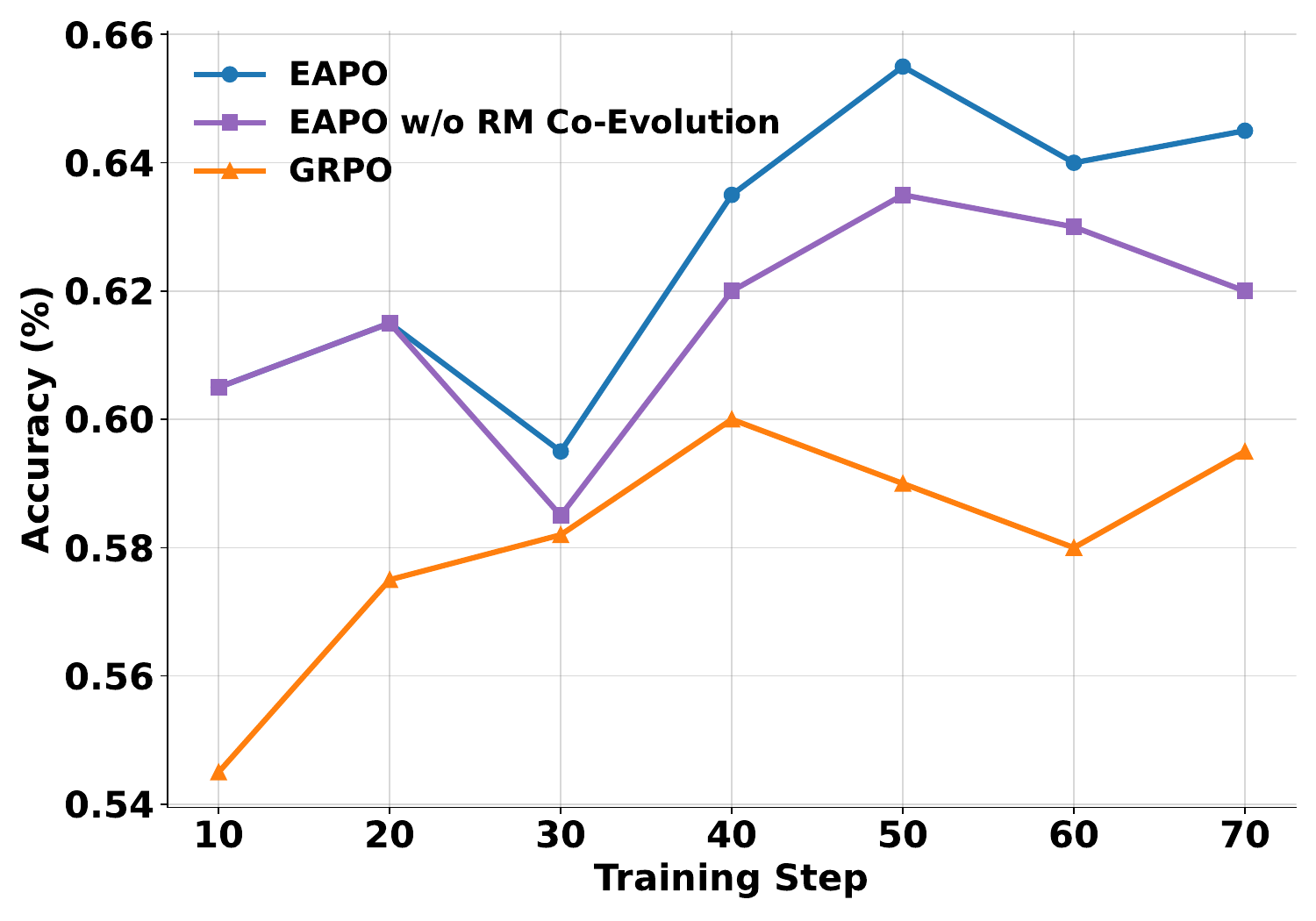} 
    \caption{Task performance on the MuSiQue in LongBench. EAPO achieves superior convergence compared to the static reward baseline and the outcome-only GRPO.}
    \label{fig:musique_score}
\end{figure}

\begin{figure}[t]
    \centering
    \includegraphics[width=0.99 \columnwidth]{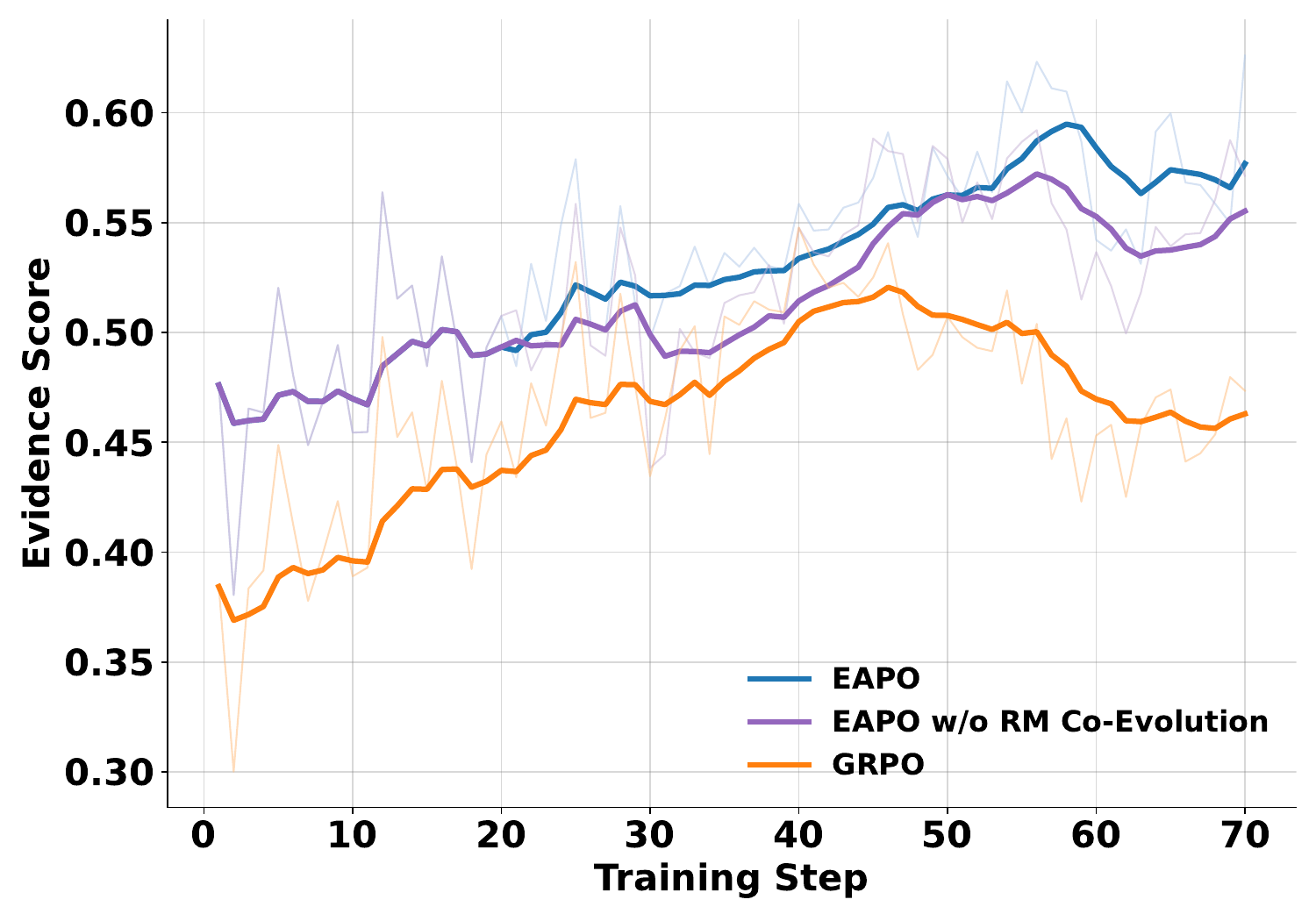} 
    \caption{Evidence Quality scores during training. EAPO explicitly optimizes the retrieval process, leading to consistently higher-quality evidence extraction compared to the outcome-based GRPO baseline.}
    \label{fig:evidence_score}
\end{figure}

To validate EAPO, we compare it against the GRPO outcome reward baseline and a static reward model variant (w/o RM Co-Evolution) in Figure \ref{fig:musique_score} and \ref{fig:evidence_score}.

\paragraph{Faster Convergence and Higher Ceiling.}
Figure \ref{fig:musique_score} shows that EAPO achieves superior sample efficiency, rapidly outpacing GRPO. This confirms that dense process supervision provides clearer gradient signals than sparse outcome rewards. Crucially, while the static variant saturates after step 50, the full EAPO framework driven by Co-Evolution, sustains its upward trajectory, achieving a higher performance ceiling.

\paragraph{Evidence Supervision Enhances Retrieval.}
As shown in Figure \ref{fig:evidence_score}, GRPO's implicit optimization of evidence is inefficient. In contrast, EAPO maintains a substantial and consistent lead in evidence scores by explicitly targeting evidence utility. This validates that the dense reward mechanism acts as a critical accelerator, ensuring the model prioritizes high-fidelity retrieval from the outset. To rigorously validate that these model-based score improvements reflect true evidence quality, we conducted a targeted human verification study. As detailed in Appendix \ref{sec:human_eval}, human annotators confirmed that 97.3\% of the evidence extracted by EAPO strictly supports the final answer, outperforming the GRPO baseline (92.6\%).

\subsection{Reward Model Analysis}
\begin{figure}[t]
    \centering
    \includegraphics[width=0.93 \columnwidth]{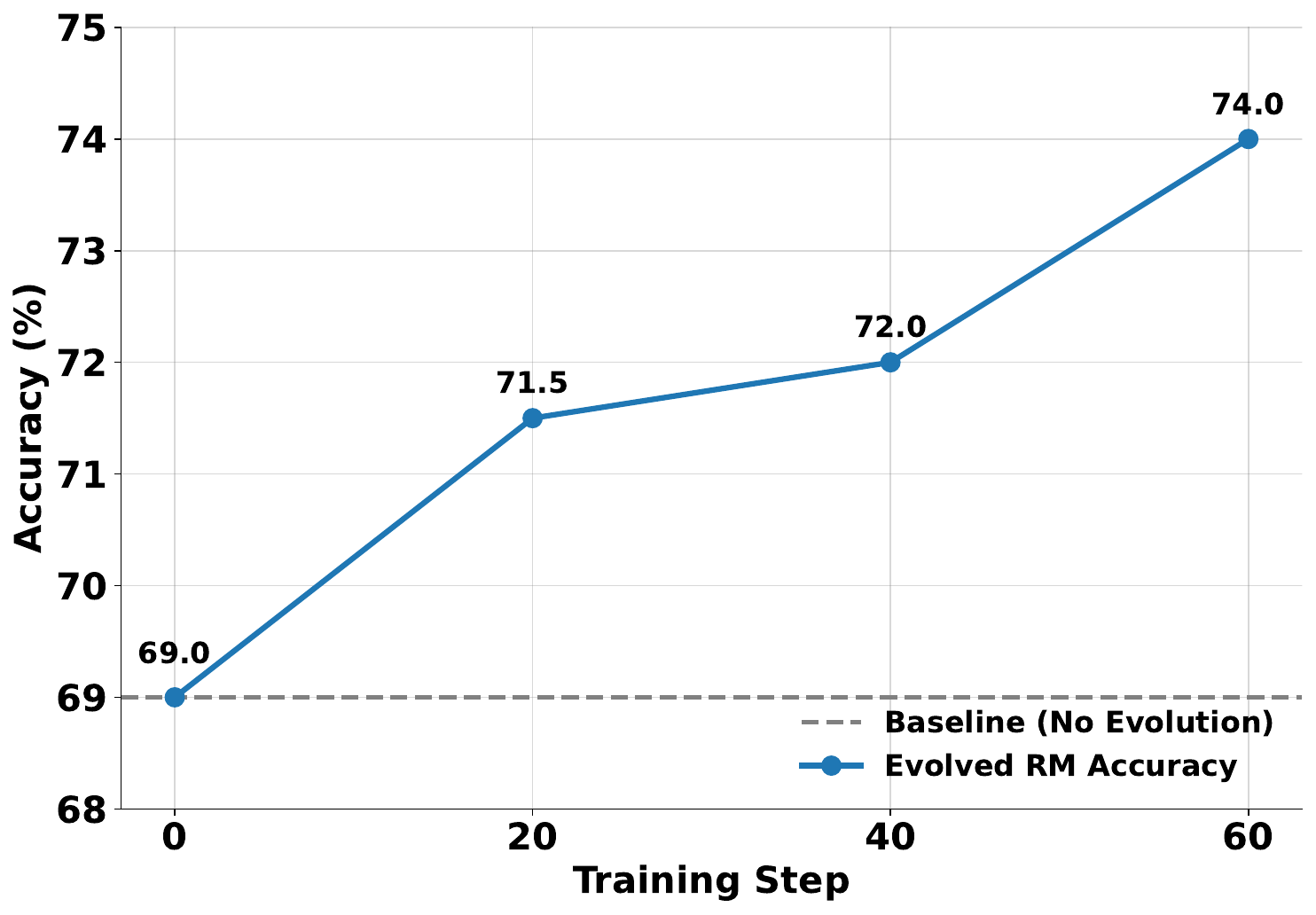}
    \caption{Evolution of RM Best-of-N Selection Accuracy ($N=6$).}
    \label{fig:rm_accuracy}
\end{figure}

To quantify the gain of Reward Model, we evaluate the RM's Best-of-N Selection Accuracy ($N=6$) followed by \citet{DBLP:journals/corr/abs-2504-02495}. Using a set of samples with grounded evidence constructed in Section 2.2, we define the Oracle optimum among 6 sampled trajectories as the one maximizing ROUGE-L Recall. Accuracy is measured by the frequency with which the RM's top-ranked candidate matches this Oracle. As shown in Figure \ref{fig:rm_accuracy}, the accuracy improves from 69.0\% to \textbf{74.0\%} over 60 steps, validating that co-evolution effectively aligns the RM with high-fidelity evidence quality.

\subsection{Hyperparameter Experiment}

\begin{figure}[t]
    \centering
    \includegraphics[width=0.99 \columnwidth]{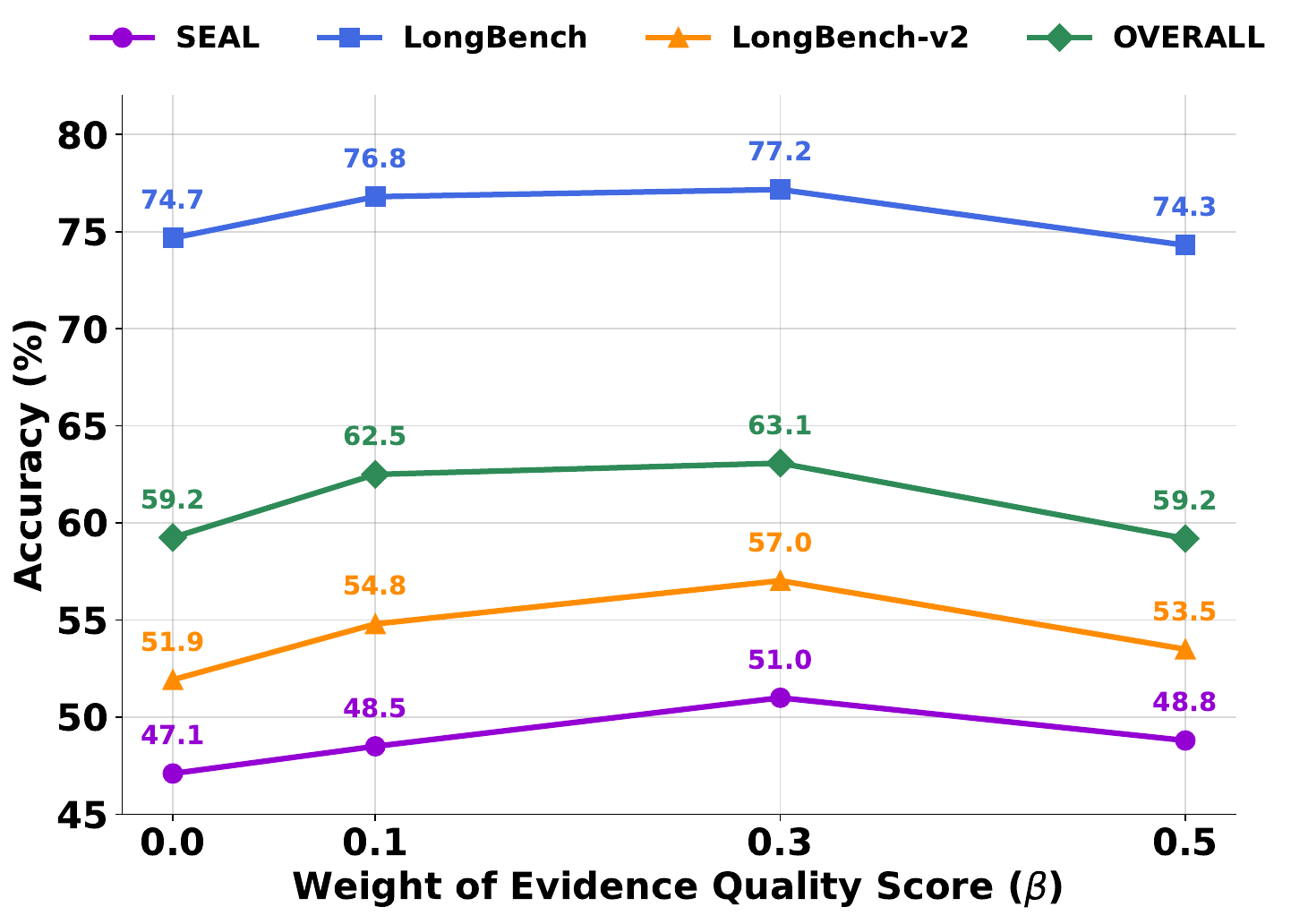}
    \caption{Comparison of different evidence quality reward weights.}
    \label{beta_ablation}
\end{figure}
We analyze the impact of $\beta$, which balances the evidence reward ($R_e$) against the outcome reward ($0.9 - \beta$). As shown in Figure \ref{beta_ablation}, increasing $\beta$ from 0.0 (outcome-only) to 0.3 yields consistent gains across all benchmarks, with Overall accuracy peaking at 63.1\%. This confirms the value of dense evidence supervision. However, further increasing $\beta$ to 0.5 causes a sharp decline back to 59.2\%, suggesting that over-weighting intermediate steps distracts from the final answer. Thus, $\beta = 0.3$ offers the optimal trade-off between process and outcome supervision.

\subsection{Error Analysis}

\begin{figure}[t]
    \centering
    \includegraphics[width=0.99\columnwidth]{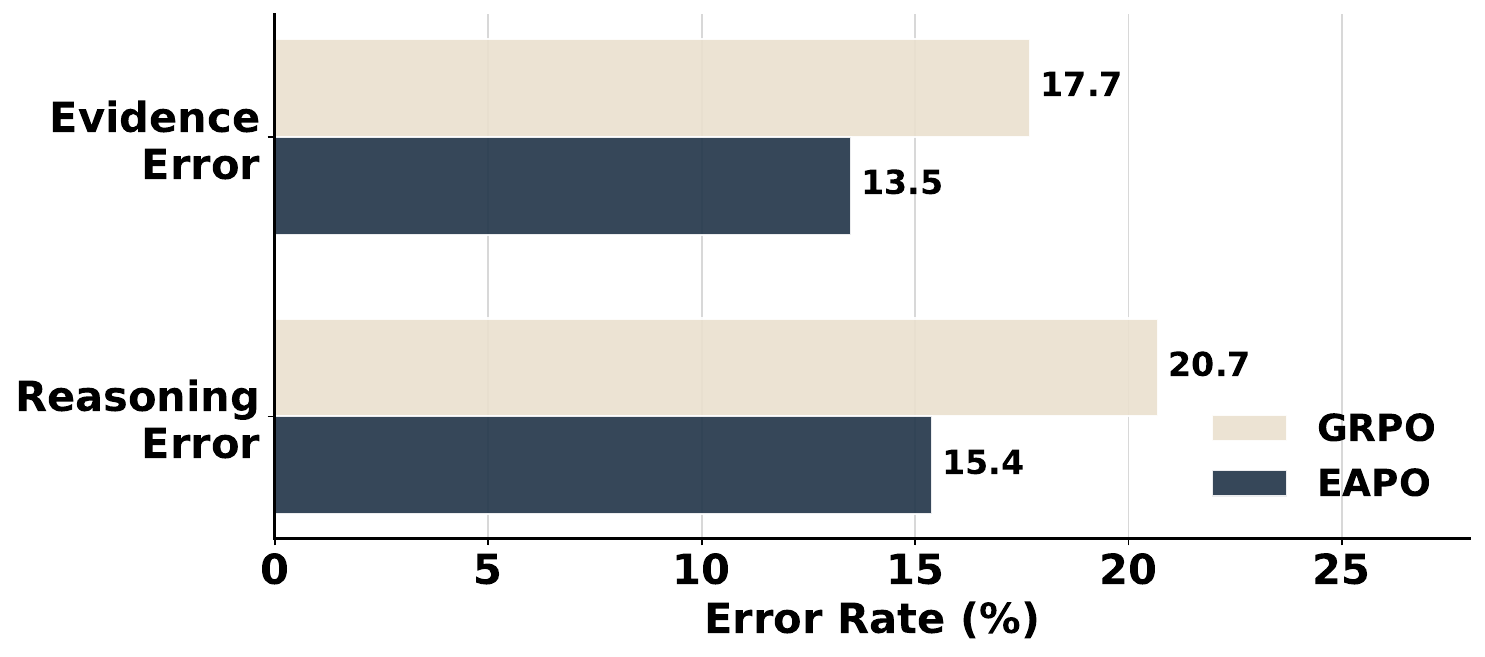}
    \caption{Error analysis on LongBench-v2. The comparison focuses on the reduction of Evidence and Reasoning errors achieved by EAPO relative to the GRPO baseline.}
    \label{fig:error_analysis_bar}
\end{figure}

We conducted a fine-grained error analysis on LongBench-v2, centering our investigation on two critical failure modes: Evidence Error (failed or partial retrieval) and Reasoning Error (correct evidence but flawed logic). 
As visualized in Figure \ref{fig:error_analysis_bar}, EAPO demonstrates a simultaneous reduction in both highlighted categories: Evidence Errors decline from 17.7\% to 13.5\%, and Reasoning Errors from 20.7\% to 15.4\%. 
This trend validates our core hypothesis that dense process supervision not only minimizes retrieval failures but also generates a cascading positive effect, grounding subsequent reasoning in higher-fidelity information to mitigate downstream logic errors.

\section{Related Work}
\subsection{Long-Context Reasoning}
Long-context reasoning is a prominent area of focus in the development of LLMs, as the context window determines the upper limit of a model's single-pass reasoning capabilities. Mainstream approaches generally fall into four categories: 1) Retrieval-Augmented Generation (RAG) \cite{DBLP:conf/iclr/0008PWM0LSBSC24, DBLP:conf/acl/JinLDZZWLQD25}, which relies on external retrievers that may become bottlenecks; 2) Agent-based Frameworks \cite{DBLP:conf/nips/Zhang0CPZA24, DBLP:journals/corr/abs-2509-06436}, which decompose contexts via multiple agents but incur high computational latency; and 3) Architectural Modifications \cite{DBLP:conf/iclr/ChenQTLL0J24, DBLP:conf/icml/DingZZXSX0Y24}, which often struggle with complexity constraints. Distinct from these patches, our research focuses on 4) Reinforcement Learning (RL). Recent long-context RL methods primarily rely on sparse outcome signals or data enhancements. For instance, LongReward \cite{DBLP:conf/acl/ZhangHLCHNHDFL25} optimizes automated outcome metrics via DPO, LoongRL \cite{DBLP:journals/corr/abs-2510-19363} synthesizes complex reasoning data for standard GRPO, and QwenLong-l1 \cite{DBLP:journals/corr/abs-2505-17667} mitigates instability in outcome-based training. While effective, they inherently struggle with ungrounded reasoning. EAPO distinguishes itself by injecting dense, process-oriented supervision directly into the RL framework to explicitly tackle this challenge.

\subsection{Reinforcement Learning and PRMs}
RL has proven effective for enhancing LLM reasoning \cite{DBLP:journals/corr/dpsk, DBLP:journals/corr/abs-2503-09516}. Beyond simple outcome-based rewards, Process-based Reward Models (PRMs) \cite{DBLP:journals/corr/abs-2305-20050, DBLP:journals/corr/abs-2504-16828} provide more granular, step-by-step guidance. However, recent PRM advancements remain heavily concentrated in structured domains like mathematics (e.g., Qwen2.5-Math \cite{DBLP:conf/acl/ZhangZWZLYLZL25}, DeepSeekMath-v2 \cite{shao2025deepseekmath}, LSRL \cite{DBLP:conf/emnlp/Ren25}) and code generation (e.g., Posterior-GRPO \cite{DBLP:journals/corr/abs-2508-05170}), which inherently allow for rigorous logical verification. Because long-context reasoning lacks such strict linear verifiability, PRMs have been historically absent from this field. EAPO bridges this critical gap. Driven by our empirical finding that evidence extraction is the decisive bottleneck, we pioneer the PRM paradigm in long-context reasoning by isolating and explicitly rewarding this specific extraction process. Furthermore, we introduce an Adaptive Reward-Policy Co-Evolution mechanism to dynamically maintain PRM accuracy, effectively preventing reward degradation during training.

\section{Conclusion}
In this work, we introduced Evidence-Augmented Policy Optimization (EAPO) to overcome the limitations of sparse outcome rewards in long-context reasoning. Anchored in the Evidence-Augmented Reasoning paradigm, EAPO shifts the optimization focus from final results to process quality, utilizing a Reward Model to provide dense Group-Relative Evidence Rewards. Crucially, to sustain rigorous supervision, we integrated an Adaptive Reward-Policy Co-Evolution mechanism, which iteratively sharpens the reward model's discriminative capability in tandem with the policy. Extensive evaluations demonstrate that EAPO significantly outperforms state-of-the-art baselines, paving the way for trustworthy, precision-centric autonomous agents.

\bibliography{custom}

\appendix

\section{Experiment Details}
\label{appendix:exp-detail}

\paragraph{Implementation Details}
We implement our reinforcement learning pipeline based on the VeRL framework \citep{verl}. For rollout generation, we set the sampling temperature to $1.0$, with a maximum context window of 120k tokens for input and 8k tokens for output generation. The group size is configured as $G=6$. Optimization is performed with a global batch size of 64 and a mini-batch size of 32, utilizing a constant learning rate of $2 \times 10^{-6}$. All experiments are executed on a computational cluster equipped with 16 NVIDIA H20 GPUs (90GB VRAM). The RL training time for 30B is approximately 72 GPU hours, and for 14B it is approximately 40 GPU hours. The SFT training for reward model is approximately 2 GPU hours.

\paragraph{Wikipedia-based Mixed QA Construction}
To detail the \textit{Wikipedia-based Mixed QA} subset mentioned in the paper, we prioritize natural context continuity by sourcing Wikipedia articles exceeding 64k tokens. Through HTML parsing, we separate content into prose (unstructured) and tables (structured) to generate diverse reasoning tasks:
\begin{itemize}
    \item \textbf{Structured QA (Single-Fragment):} Focuses on analytical reasoning over discrete table segments. These tasks require the model to perform filtering, ranking, and numerical calculation within a single structured evidence block.
    \item \textbf{Heterogeneous Mixed QA (Multi-Fragment):} Designed to bridge structured and unstructured modalities. This category includes \textit{Text-and-Table QA}, which demands synthesizing information from free-text prose and tabular data, and \textit{Multi-Table QA}, which requires linking logic across multiple interrelated tables.
\end{itemize}

\section{Analysis of Reasoning Diversity}
\label{sec:reasoning_diversity}

In Section \ref{pre-exp}, we observed that optimizing solely the reasoning step (EAR+LLM-R) or adding a reasoning judge atop evidence evaluation (EAR+LLM-E+LLM-R) yields no performance improvement, and even causes a slight drop compared to evaluating evidence alone (LLM-E). To understand the underlying causes of this phenomenon, we empirically analyze the semantic diversity of the generated reasoning paths.

In our framework, evidence extraction strictly precedes reasoning execution. We hypothesize that both the content and the diversity of the downstream reasoning are heavily anchored to the extracted evidence. To verify this, we measured the semantic diversity of the reasoning content across the sampled paths for each method. We employed an LLM-as-a-Judge approach to evaluate whether the generated paths for a given query are semantically diverse or highly consistent (i.e., sharing identical core logic and conclusions). 

\begin{table}[h]
\centering
\small
\resizebox{\columnwidth}{!}{
\begin{tabular}{lcc}
\toprule
\begin{tabular}[c]{@{}l@{}}\textbf{Sampling} \\ \textbf{Method}\end{tabular} & 
\begin{tabular}[c]{@{}c@{}}\textbf{Semantically} \\ \textbf{Diverse}\end{tabular} & 
\begin{tabular}[c]{@{}c@{}}\textbf{Semantically} \\ \textbf{Consistent}\end{tabular} \\
\midrule
EAR + LLM-E         & 24.0\% & 76.0\% \\
EAR + LLM-R         & 3.5\%  & 96.5\% \\
EAR + LLM-E + LLM-R & 4.5\%  & 95.5\% \\
\bottomrule
\end{tabular}
}
\caption{Semantic diversity of reasoning outputs across different sampling configurations. When the reasoning evaluator (LLM-R) is applied without varying the evidence, the resulting reasoning paths become highly homogeneous.}
\label{tab:reasoning_diversity}
\end{table}

The results, presented in Table \ref{tab:reasoning_diversity}, explain the limited efficacy of LLM-R. When the evidence content is fixed (as in EAR+LLM-R), the subsequent reasoning content becomes almost completely static, exhibiting 96.5\% semantic consistency. Because fixing the evidence removes almost all variation in the downstream outputs, the exploration space for the reasoning step becomes severely restricted. 
Consequently, the reasoning judge is forced to evaluate candidate paths that share nearly identical reasoning content and lead to the exact same results, making the evaluation step highly redundant.

\section{Human Verification of Evidence Quality}
\label{sec:human_eval}

We conducted an additional targeted manual annotation study to explicitly verify how often the extracted evidence truly supports the final answer under strict human review.

To maintain query consistency and ensure a fair comparison, we first identified the intersection subset of queries that were answered correctly by both the EAPO model and the GRPO baseline. From this exact intersection, we randomly sampled 100 queries and extracted their complete reasoning trajectories from both models. 

We recruited three human annotators with NLP research backgrounds to perform a blind evaluation of these samples. They were tasked with answering a strict, boolean question: \textit{"Does the information within the evidence block alone provide sufficient and logically sound support to derive the final answer?"}

The human verification results demonstrated that, on average, 97.3\% of the EAPO samples were judged as having evidence that completely supports the final conclusion, compared to an average of 92.6\% for the GRPO baseline. This direct assessment corroborates our automated metrics, clearly indicating that the EAPO training paradigm significantly enhances the model's ability to accurately cite supporting evidence. It ensures that the generated reasoning is not just outcome-correct, but strictly grounded and human-verifiable.

\section{Needle in a Haystack Retrieval Experiment}
\begin{figure}[t]
    \centering
    
    \includegraphics[width=0.99 \columnwidth]{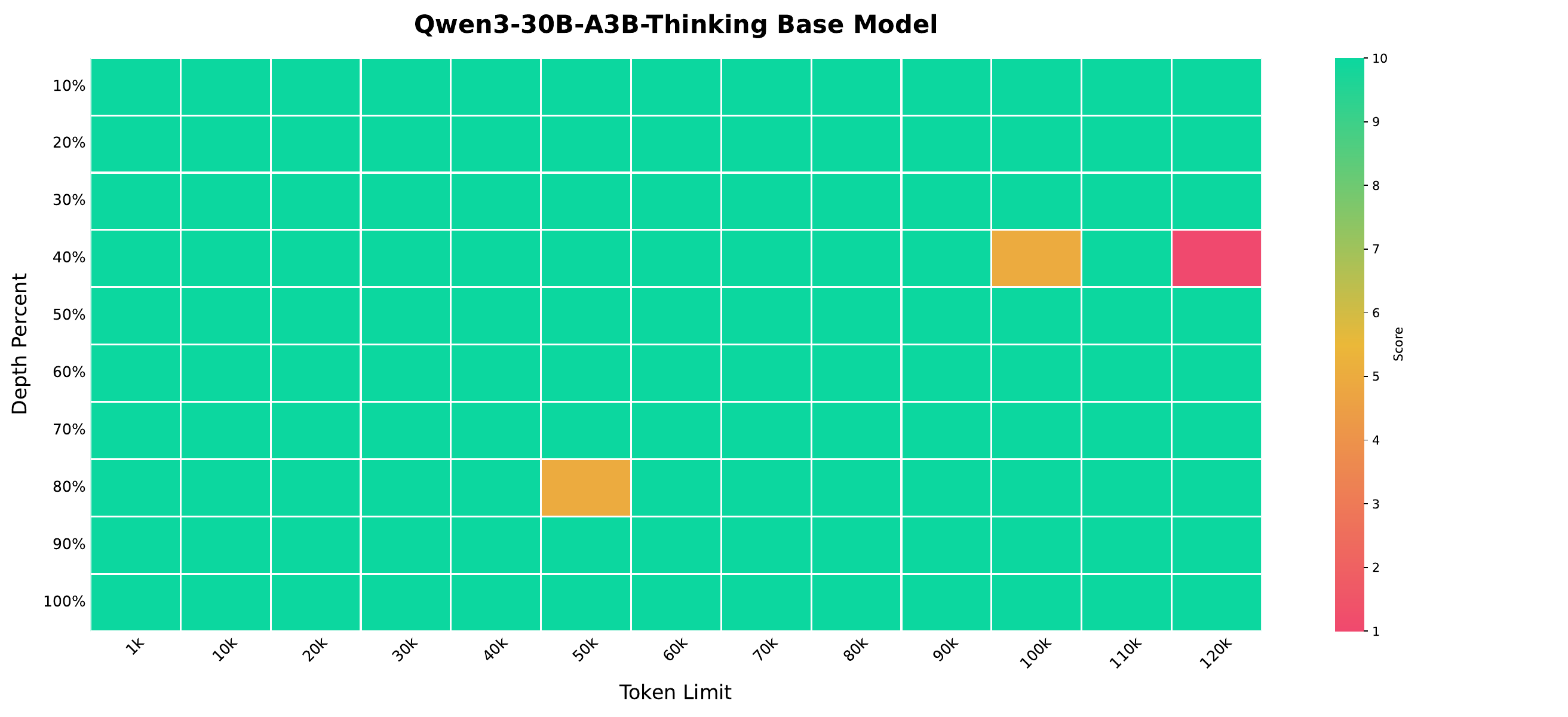}
    \vspace{-15pt} 
    
    \vspace{5pt}   

    \includegraphics[width=0.99 \columnwidth]{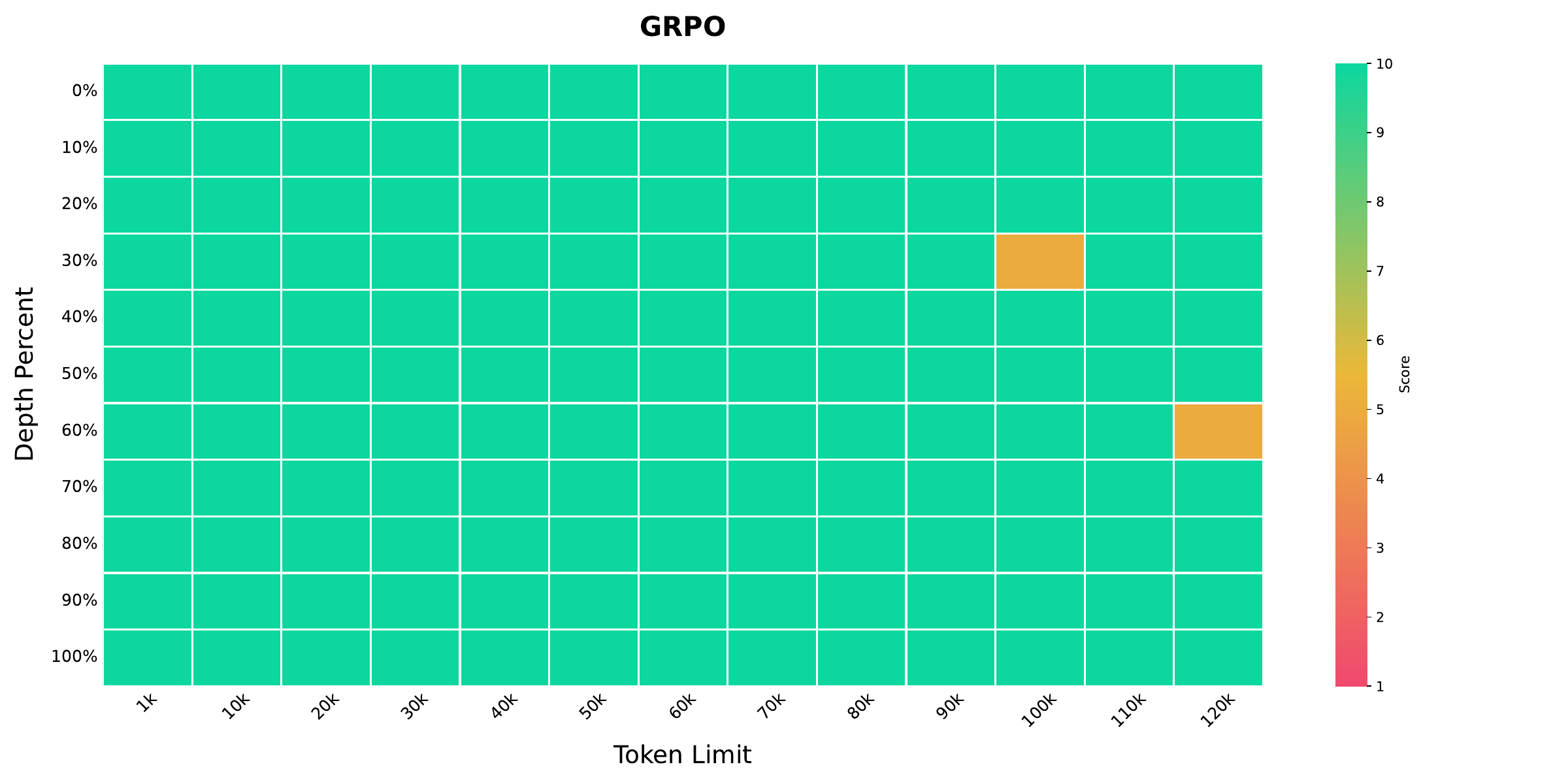}
    \vspace{-15pt}
    \vspace{5pt}

    \includegraphics[width=0.99 \columnwidth]{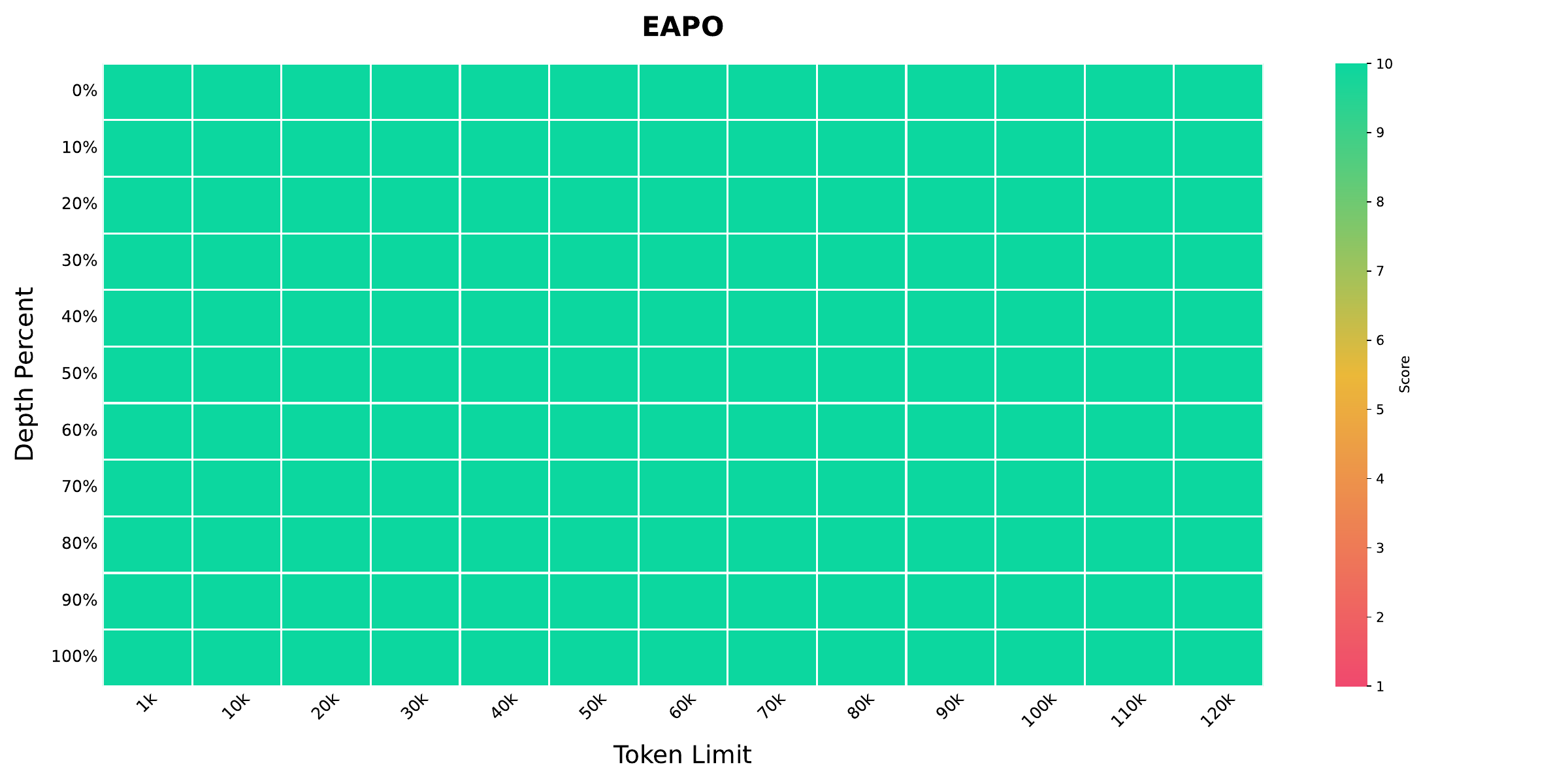}
    \vspace{-15pt}
    
    \caption{Needle in a Haystack retrieval performance across varying context lengths and depths.}
    \label{needle}
\end{figure}

We additionally evaluate retrieval performance using the "Needle in a Haystack" (NIAH) benchmark \cite{needle}. As shown in Figure \ref{needle}, both the \textbf{Base Model} and \textbf{GRPO} fall slightly short of perfection, failing to retrieve needles in a few specific instances. In contrast, \textbf{EAPO} achieves a flawless 100\% retrieval rate. The "all-green" heatmap confirms that our dense evidence supervision further refines retrieval robustness, ensuring precise information localization across the entire context window.

\section{Detailed Prompt} \label{prompt}

We explicitly map each prompt to its corresponding module across our preliminary experiments (Section 2) and the EAPO framework (Section 3):

\begin{itemize}
    \item \textbf{Prompt for EAR (Generation Prompt):} Utilized for generating reasoning paths in both the standard EAR baseline (Section 2) and the EAPO policy generation phase (Section 3).
    \item \textbf{Prompt for Evidence Evaluation:} Employed by the evidence judge (LLM-E) in the preliminary study (Section 2.2) to score evidence utility, and also used by the Reward Model to calculate the Group-Relative Evidence Quality in EAPO (Section 3.2).
    \item \textbf{Prompt for Reasoning Evaluation:} Used specifically by the reasoning judge (LLM-R) in the preliminary study (Section 2.2) to assess the logic and coherence of the generated reasoning steps.
    \item \textbf{Prompt for Answer Evaluation (Outcome Evaluation):} Applied to compute the Outcome Accuracy Reward in EAPO (Section 3.2) and utilized for the final answer accuracy assessment in our main evaluation (Section 4).
\end{itemize}

\lstset{
    basicstyle=\small\ttfamily,
    breaklines=true,            
    breakatwhitespace=true,    
    columns=fullflexible,       
    frame=none,          
    aboveskip=0.5em,
    belowskip=0.5em
}

\newtcolorbox{promptbox}[1][]{
  colback=gray!5!white,
  colframe=gray!75!black,
  title={Prompt},
  fonttitle=\bfseries,
  coltitle=white,
  enhanced,
  attach boxed title to top left={yshift=-2mm, xshift=2mm},
  boxed title style={colback=gray!80!black},
  #1
}

\begin{promptbox}[title= Prompt for EAR, breakable]
    \small 
    Please read the following content and answer the question below.
    
    Content:
    
    \{content\}

    Question:
    
    \{question\}
    
    \vspace{0.5em}
    You must strictly adhere to the following four-part structure for your entire output. Do not include any introductory phrases or text outside of this format.
    \vspace{0.5em}
    Required Output Format:
    
    <analysis>
    
    [Articulate your analysis of the question, identify the specific information required, and outline your strategy for locating and extracting relevant evidence from the provided content. Do not answer the question directly.]
    
    </analysis>
    
    <evidence>
    
    [Insert all verbatim, unmodified quotes from the content that directly support the final answer. Each piece of evidence should be on a new line and be as comprehensive as possible, using '...' to link related but separate text fragments if necessary.]
    
    </evidence>
    
    <reasoning>
    
    [Provide a step-by-step logical explanation that connects the extracted evidence to the final answer. Clearly demonstrate how the evidence leads to your conclusion.]
    
    </reasoning>
    
    <answer>
    
    [State the final, concise answer. This answer must be derived solely from the evidence provided above.]
    
    </answer>
\end{promptbox}

\begin{promptbox}[title= Prompt for Evidence Evaluation, breakable]
\label{evidence prompt}
    \small 

    You are a master strategist and information architect.
    Objective: Critically evaluate competing sets of `Evidence`. Your mission is to act as a strategic filter, identifying the single most valuable set of evidence for constructing a high-quality, comprehensive, and accurate answer to the `Original Question`.
    
    Inputs:
    
      * Original Question:
      
      \{question\}
      
      * Evidence Choices:
      
      \{choices\}
      
    \vspace{0.5em}
    Core Evaluation Principle: Comparative Analysis
    Your evaluation must be relational, not absolute. Do not assess each choice in isolation. The score for any given choice must be determined by comparing its performance against the other available options across the criteria below. For a choice to earn a top score, it must be demonstrably superior *relative to its competitors*.
    
    Evaluation Criteria:
    You must use these criteria as the dimensions for your comparison:
    
    1.  Relevance \& Focus: How directly does the evidence address all facets of the `Original Question`? Is it tightly focused, or diluted with irrelevant "noise"?
    
    2.  Depth \& Breadth: Does the evidence provide comprehensive, multi-faceted coverage of the topic? Or is 
    it superficial, one-sided, or incomplete?
    
    3.  Utility \& Specificity: Is the evidence composed of concrete, factual, and directly usable information? Or is it vague, overly general, and requires heavy interpretation?
    
    4.  Coherence: Are the individual pieces of evidence within the set consistent with each other? Or do internal contradictions create ambiguity and undermine reliability?
    
    Scoring Rubric (1-5 Scale):
    You must assign a score to each choice based on the following strict rubric. Remember to apply these definitions through the lens of the Core Evaluation Principle.
    
    *   1: Unusable: The evidence is fundamentally flawed. It is irrelevant, internally contradictory, or so vague that it provides no usable basis for an answer.
    
    *   2: Poor: The evidence has major deficiencies. It may be only tangentially relevant, highly imbalanced, or lacking specific facts, making it a weak foundation, especially when compared to stronger options.
    
    *   3: Adequate: The evidence is usable but clearly inferior to better options. It addresses the core question but has significant comparative gaps in depth, breadth, or specificity.
    
    *   4: Good: The evidence is a strong contender. It is highly relevant and specific, but may be slightly less comprehensive or focused than the top choice. It provides a solid foundation for a high-quality answer.
    
    *   5: Excellent: The evidence is perfectly tailored and clearly superior to all other options. It is the most comprehensive, balanced, specific, and coherent dataset available, with virtually no noise.

    \vspace{0.5em}
    
    Output Requirements:
    Your entire output must be a single, valid JSON object structured exactly as follows, containing three keys: `reason`, `scores`, and `best\_choice`.

    \vspace{0.5em}
    \{{
    
        "reason": "A string containing your detailed comparative analysis. Your analysis must: 1. Justify each score by explicitly comparing that choice to the others on the key `Evaluation Criteria` (e.g., 'Choice A earns a 4 for Depth because it is more comprehensive than Choice B, but Choice C covers one additional aspect, making it superior'). 2. Use the language of the `Scoring Rubric` to anchor your comparative judgments. 3. Conclude with a definitive statement explaining why the `best\_choice` is superior *relative to the alternatives*.",
        
        "scores": [An array of integers, where each integer is the 1-5 score corresponding to each choice in the provided order.],
        
        "best\_choice": An integer representing the 0-based index of the single best evidence choice.
    
    }\}

\end{promptbox}

\begin{promptbox}[title= Prompt for Answer Evaluation, breakable]
    \small 
    
    Your job is to look at a question, a gold target, and a predicted answer, and then assign a grade of either ["CORRECT", "INCORRECT", "NOT\_ATTEMPTED"].
    First, I will give examples of each grade, and then you will grade a new example.

    The following are examples of CORRECT predicted answers.
    
    Question: What are the names of Barack Obama's children?
    
    Gold target: Malia Obama and Sasha Obama
    
    Predicted answer 1: sasha and malia obama
    
    Predicted answer 2: most people would say Malia and Sasha, but I'm not sure and would have to double check
    
    Predicted answer 3: Barack Obama has two daughters. Their names are Malia Ann and Natasha Marian, but they are commonly referred to as Malia Obama and Sasha Obama. Malia was born on July 4, 1998, and Sasha was born on June 10, 2001.
    
    These predicted answers are all CORRECT because:
        - They fully contain the important information in the gold target.
        - They do not contain any information that contradicts the gold target.
        - Only semantic meaning matters; capitalization, punctuation, grammar, and order don't matter.
        - Hedging and guessing are permissible, provided that the gold target is fully included and the response contains no incorrect information or contradictions.

    The following are examples of INCORRECT predicted answers.
   
    Question: What are the names of Barack Obama's children?
    
    Gold target: Malia and Sasha
    
    Predicted answer 1: Malia.
    
    Predicted answer 2: Malia, Sasha, and Susan.
    
    Predicted answer 3: Barack Obama does not have any children.
    
    Predicted answer 4: I think it's either Malia and Sasha. Or it could be Malia and Jackie. Or it could be Joey and Malia.
    
    Predicted answer 5: It's possible you may mean Betsy and Olivia. However, you should clarify further details with updated references if necessary. Is that the correct answer?
    
    Predicted answer 6: It may be the case that Obama's child is named James. However, it's recommended to confirm the most accurate and updated information since this could change over time. This model may not always reflect the most current information.
    
    These predicted answers are all INCORRECT because:
        - A factual statement in the answer contradicts the gold target. Incorrect statements that have some hedging (e.g., "it is possible that", "although i'm not sure, i think") are also considered incorrect.

    The following are examples of NOT\_ATTEMPTED predicted answers.
    
    Question: What are the names of Barack Obama's children?
    
    Gold target: Malia and Sasha
    
    Predicted answer 1: I don't know.
    
    Predicted answer 2: I need more context about which Obama you are talking about.
    
    Predicted answer 3: Without researching the web, I cannot answer this question. However, I can tell you that Barack Obama has two children.
    
    Predicted answer 4: Barack Obama has two children. I know that one of them is Malia, but I'm not sure about the other one.
    
    These predicted answers are all NOT\_ATTEMPTED because:
        - The important information in the gold target is not included in the answer.
        - No statements in the answer contradict the gold target.

    Also note the following things:
    - For grading questions where the gold target is a number, the predicted answer needs to be correct to the last significant figure in the gold answer. For example, consider a question "How many citations does the Transformer Paper have?" with gold target "120k". 
    
        - Predicted answers "120k", "124k", and 115k" are all CORRECT. 
        
        - Predicted answers "100k" and "113k" are INCORRECT. 
        
        - Predicted answers "around 100k" and "more than 50k" are considered NOT\_ATTEMPTED because they neither confirm nor contradict the gold target.
        
    - The gold target may contain more information than the question. In such cases, the predicted answer only needs to contain the information that is in the question.
    
        - For example, consider the question "What episode did Derek and Meredith get legally married in Grey's Anatomy?" with gold target "Season 7, Episode 20: White Wedding". Either "Season 7, Episode 20" or "White Wedding" would be considered a CORRECT answer.
        
    - Do not punish predicted answers if they omit information that would be clearly inferred from the question.
        - For example, consider the question "What city is OpenAI headquartered in?" and the gold target "San Francisco, California". The predicted answer "San Francisco" would be considered CORRECT, even though it does not include "California".
        
        - Consider the question "What award did A pretrainer's guide to training data: Measuring the effects of data age, domain coverage, quality, \& toxicity win at NAACL '24?", the gold target is "Outstanding Paper Award". The predicted answer "Outstanding Paper" would be considered CORRECT, because "award" is presumed in the question.
        
        - For the question "What is the height of Jason Wei in meters?", the gold target is "1.73 m". The predicted answer "1.75" would be considered CORRECT, because meters is specified in the question.
        
        - For the question "What is the name of Barack Obama's wife?", the gold target is "Michelle Obama". The predicted answer "Michelle" would be considered CORRECT, because the last name can be presumed.
        
    - Do not punish for typos in people's name if it's clearly the same name. 
        - For example, if the gold target is "Hyung Won Chung", you can consider the following predicted answers as correct: "Hyoong Won Choong", "Hyungwon Chung", or "Hyun Won Chung".
    
    \vspace{0.5em}
    Here is a new example. Simply reply with either CORRECT, INCORRECT, NOT ATTEMPTED. Don't apologize or correct yourself if there was a mistake; we are just trying to grade the answer.
    
    Question: {question}
    
    Gold target: {correct\_answer}
    
    Predicted answer: {response}

    Grade the predicted answer of this new question as one of:
    
    A: CORRECT
    
    B: INCORRECT
    
    C: NOT\_ATTEMPTED
    
    Just return the letters "A", "B", or "C", with no text around it. 
    
\end{promptbox}

\begin{promptbox}[title= Prompt for Reasoning Evaluation, breakable]
\label{reasoning prompt}
    \small 
    You are a Strategic Logic Evaluator and Reasoning Analyst.
    
    **Objective:** Evaluate a set of candidate outputs representing the **NEXT STEP** in a reasoning chain. You must determine which candidate represents the most logical, accurate, and valuable progression towards answering the `Original Question`, given the `Historical Reasoning Context`.
    
    **Inputs:**
    * **Original Question:**
    {question}
    
    * **Historical Reasoning Context (The Unfinished Process):**
    {approved\_reasoning\_process}
    
    * **Candidate Next Steps:**
    {choices}
    
    **Core Evaluation Principles:**
    Assess each choice based on its logical validity as a continuation of the provided context.
    
    1.  **Logical Validity \& Advancement:**
        * **Continuation:** If the `Historical Reasoning Context` is sound, the candidate should logically follow it, deepening the analysis or moving to the next deduction.
        * **Correction/Pivot:** If the `Historical Reasoning Context` contains flaws, contradictions, or weak assumptions, a candidate that **identifies and corrects** these issues is superior to one that blindly follows a wrong path.
        * **Flexibility:** The candidate is NOT required to strictly adhere to the *conclusions* of the history, but it must strictly adhere to the *facts/evidence* cited within it (no external hallucinations).
    
    2.  **Evidence Grounding:** While the logical path can change, facts cannot be invented. The candidate must still rely on the factual evidence provided within the context (or implicitly known general knowledge if applicable to the domain), without hallucinating new specific data.
    
    3.  **Efficiency:** Does this step actually move the needle? Avoid circular reasoning or restating what has already been established in the history.
    
    **Scoring Rubric (1-4 Scale):**
    
    * **1: Critical Failure (Hallucination/Illogical Jump):**
        * The output invents facts not present in the context.
        * The output makes a logical leap that makes no sense, neither following the history nor offering a valid correction.
        * The output completely ignores the `Original Question`.
    
    * **2: Poor (Redundant/Weak):**
        * The output merely restates the `Historical Reasoning Context` without adding new value.
        * The output blindly follows a clearly erroneous path established in the history (fails to critical thinking).
        * The logic is circular or trivial.
    
    * **3: Good (Valid Progression):**
        * A solid, logical next step. It accepts the context and moves forward reasonably.
        * If the history is sound, this choice extends it correctly.
        * No hallucinations; safe and compliant.
    
    * **4: Excellent (Insightful Advancement or Critical Correction):**
        * The output demonstrates superior reasoning.
        * It might provide a **necessary pivot** if the previous logic was heading in the wrong direction.
        * It synthesizes previous points to reach a breakthrough or a highly distinct intermediate conclusion.
        * It maximizes the probability of correctly answering the `Original Question`.
    
    **Output Requirements:**
    You must output a **single, valid JSON object** containing the following keys. Do not include markdown formatting or any text outside the JSON object.
    
    \{{
    
      "reason": "A detailed analysis string. You must: 1) Evaluate the state of the 'Historical Reasoning Context' (is it on the right track?). 2) Analyze each choice's logical contribution (is it a continuation or a pivot?). 3) Justify why the 'best\_choice' is the most valuable next step towards the solution.",
      "scores": [An array of integers, where each integer is the 1-4 score corresponding to each choice in the provided order.],
      "best\_choice": An integer (0-based index of the best choice)
      
    }\}
    
\end{promptbox}

\end{document}